\documentclass[preprint,12pt]{elsarticle}

\usepackage{amssymb}
\usepackage{amsmath}
\usepackage{hyperref}
\usepackage{multirow} 
\usepackage{booktabs}
\usepackage{xcolor}
\usepackage{array}
\usepackage{rotating}

\newcommand{\valstd}[2]{$#1 {\scriptstyle \,\pm\, #2}$}
\newcommand{\valstdb}[2]{$\mathbf{#1} {\scriptstyle \,\pm\, #2}$}
\newcommand{\valstdu}[2]{$\underline{#1} {\scriptstyle \,\pm\, #2}$}

\begin{document}

\begin{frontmatter}



\title{Mind the Missing: Variable-Aware Representation Learning for Irregular EHR Time Series using Large Language Models}


\author[inst1]{Jeong Eul Kwon}\ead{handmadeoutlier@korea.ac.kr}
\author[inst2]{Joo Heung Yoon, MD, PhD\corref{cor1}}\ead{yoonjh@upmc.edu}

\author[inst1]{Hyo Kyung Lee, PhD\corref{cor1}}\ead{hyokyunglee@korea.ac.kr}

\cortext[cor1]{Corresponding author(s)}
\affiliation[inst1]{
    organization={School of Industrial Management Engineering, Korea University},
    addressline={145 Anam-ro, Seongbuk-gu}, 
    city={Seoul},
    postcode={02841}, 
    country={Korea}
}

\affiliation[inst2]{
    organization={Division of Pulmonary, Allergy, Critical Care, and Sleep Medicine, Department of Medicine, University of Pittsburgh},
    addressline={200 Lothrop street}, 
    city={Pittsburgh, PA},
    postcode={15261}, 
    country={USA}
}
\begin{abstract}
Irregular sampling and high missingness are intrinsic challenges in modeling time series derived from electronic health records (EHRs), where clinical variables are measured at uneven intervals depending on workflow and intervention timing. To address this, we propose VITAL — a variable-aware, large language model (LLM)-based framework tailored for learning from irregularly sampled physiological time series. VITAL differentiates between two distinct types of clinical variables: vital signs, which are frequently recorded and exhibit temporal patterns, and laboratory tests, which are measured sporadically and lack temporal structure. It reprograms vital signs into the language space, enabling the LLM to capture temporal context and reason over missing values through explicit encoding. In contrast, laboratory variables are embedded either using representative summary values or a learnable [Not measured] token, depending on their availability. Extensive evaluations on the benchmark datasets from the PhysioNet demonstrate that VITAL outperforms state-of-the-art methods designed for irregular time series. Furthermore, it maintains robust performance under high levels of missigness,  which is prevalent in real-world clinical scenarios where key variables are often unavailable.
\end{abstract}




\end{frontmatter}

\section{Introduction}
\label{sec1}
Electronic Health Records (EHRs) digitally capture a wealth of patient data generated during routine clinical care. In particular, the Intensive Care Unit (ICU) is a data-rich environment due to the need for continuous, high-resolution patient monitoring. This has led to a surge of research in medical artificial intelligence (AI), with many studies leveraging publicly available EHR datasets in combination with machine learning techniques for tasks such as early warning, outcome prediction and patient stratification \cite{le2020supervised, kim2024development, komorowski2018artificial, kwon2025clear, kim2025advancing, choi2024sequential, park2024multivariate, yoon2022artificial, kang2023current}

\begin{figure}[t]
\centering
\includegraphics[width=\textwidth]{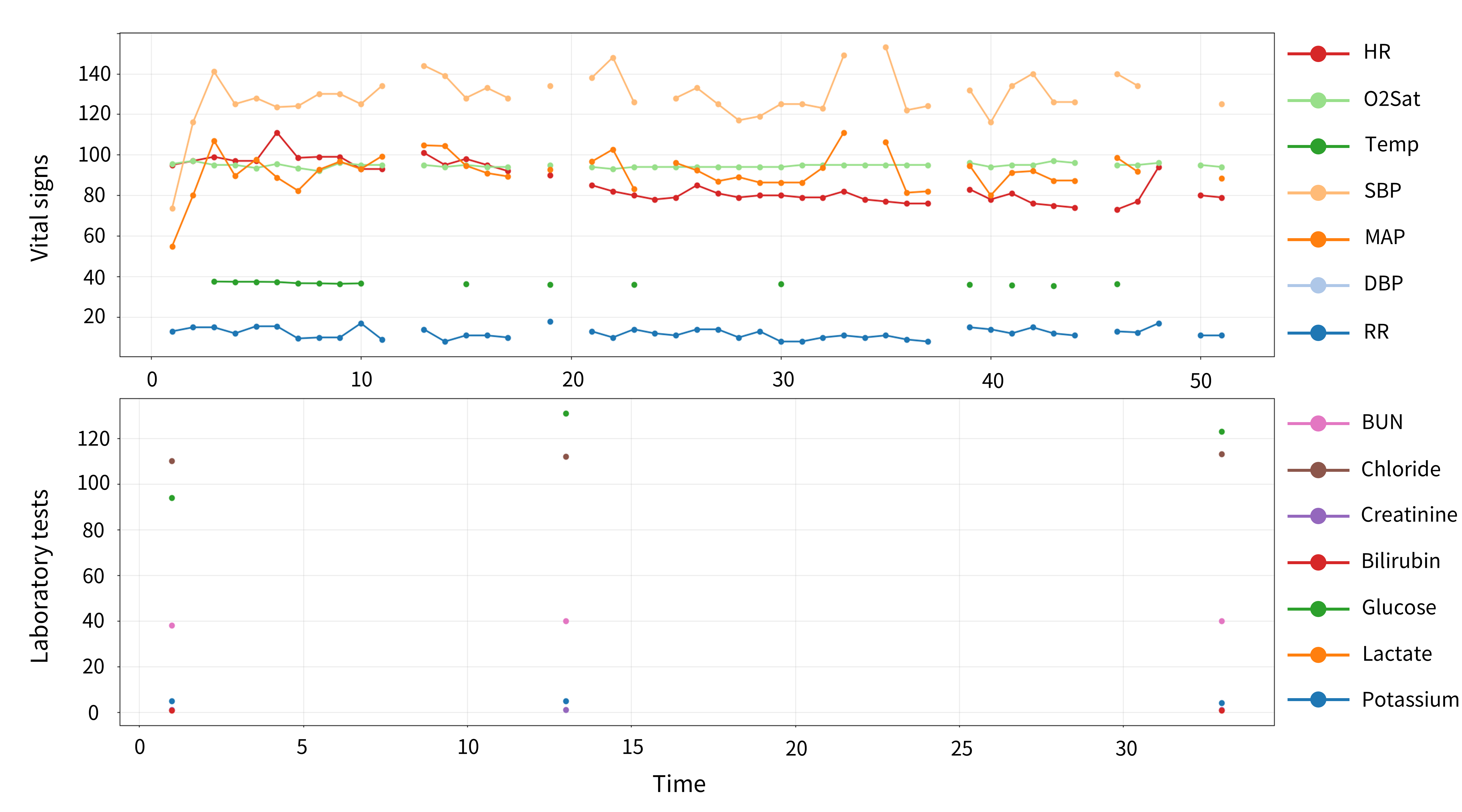}
\caption{\textbf{Irregularly sampled time series} derived from an EHR database. HR: Heart Rate, O2Sat: O2 Saturation, Temp: Temperature, SBP: Systolic Blood Pressure, MAP: Mean Arterial Pressure, DBP: Diastolic Blood Pressure, RR: Respiration Rate, BUN: Blood Urea Nitrogen. Observed values are represented as dots, and the connections between observations are shown as line.}\label{vital_lab}
\end{figure}

A common approach in these studies is to model patient records as multivariate time series, capturing the temporal evolution of physiological and clinical variables. However, in practice, EHR time series are often irregularly sampled due to variations in clinical workflows, measurement protocols, and intervention timing. As illustrated in Figure \ref{vital_lab}, such data are characterized by high missingness, irregular intervals, and unequal sequence lengths, factors that pose significant challenges for conventional time series models, which typically assume regularly spaced inputs.

This irregularity stems from differences in how clinical variables are collected. Vital signs such as heart rate, blood pressure, and respiratory rate are automatically and frequently recorded via bedside monitors, leading to regular sampling and relatively low missingness. However, even among vital signs, missingness is not uncommon due to temporary disconnections, interruptions in monitoring, or asynchronous logging. In contrast, laboratory test results such as lactate, leukocyte count, and platelets are measured manually and less frequently, often only when clinically indicated. These measurements also involve multi-step delays, including physician ordering, sample collection, lab processing, and result reporting, resulting in sparse and temporally delayed time series.

These differences create two fundamentally distinct types of variables in EHR time series: (1) frequently measured variables (vital signs), which exhibit discernible temporal patterns, and (2) infrequently measured variables (lab tests), which are sparse, irregular, and less temporally structured. Yet, many existing AI models treat all variables uniformly, often by imputing missing values and feeding the resulting time series into deep learning architectures. This uniform treatment overlooks the inherent heterogeneity of clinical data and may lead to suboptimal or misleading representations.

Recently, large language models (LLMs) have emerged as powerful general-purpose models capable of advanced reasoning, contextual understanding, and generalization across domains. Recent work  \cite{zhou2023one, chang2023llm4ts, cao2023tempo, jin2023time, sun2023test} has begun exploring LLMs for time series modeling, showing promising results that outperform traditional statistical methods and non-LLM-based approaches. These results suggest that LLMs can be effectively repurposed to handle time-series data if the inputs are appropriately restructured into a language-like format. However, applying LLMs to EHR time series presents unique challenges since not all clinical variables benefit equally from sequence modeling. Vital signs may indeed benefit from LLMs’ temporal reasoning capabilities, but sparse variables such as lab results may not. Treating all variables as dense sequences may misallocate modeling capacity and reduce robustness to missing data.

To address this, we propose VITAL (\textbf{V}ariable-aware \textbf{I}rregularly sampled \textbf{T}ime series \textbf{A}nalysis framework with \textbf{L}arge language models), a LLM-based representation learning framework designed specifically for learning from irregularly sampled physiological time series in EHR. Our approach explicitly accounts for variable-level differences in measurement frequency and temporal structure. Specifically, we seek to answer two key questions:

\begin{itemize}
  \item \textbf{Q1.} Can an LLM extract meaningful temporal contextual embeddings from vital signs, which are relatively dense and temporally structured, while explicitly handling missing observations? 
  \item \textbf{Q2.} For laboratory test variables that are highly sparse and irregularly measured, can a simpler, non-temporal embedding strategy still improve predictive performance when integrated with vital sign representations? 
\end{itemize}

As illustrated in Figure \ref{architecture}, VITAL generates temporal context-aware embeddings from irregular vital sign sequences using a pre-trained LLM, explicitly incorporating missing information. For laboratory test variables, it supplements the vital sign embedding with either representative value embeddings or a learnable [Not measured] token, depending on whether the variable was observed during the monitoring period. These representations are fused into a unified embedding for classification tasks.

The core idea of VITAL is to reprogram irregular time series into a language space that aligns with the environment of LLMs. This enables the model to transform incomplete time series into fixed-length embedding vectors effectively, without modifying any parameters of the pre-trained LLM, while fully leveraging its inherent reasoning ability.

To comprehensively evaluate the effectiveness of VITAL, we conducted comparative experiments against several state-of-the-art (SoTA) methods specifically designed for irregularly sampled time series. Using two representative healthcare datasets, P19 \cite{reyna2020early} and P12 \cite{goldberger2000physiobank}, VITAL achieved superior performance, outperforming previous SoTA models by absolute AUROC margins of 0.1\% and 0.9\%, and absolute AUPRC improvements of 4.4\% and 3.4\%, respectively. Furthermore, in the leave-fixed-sensors-out setting which simulates real-world clinical scenarios where some variables are entirely unobserved, VITAL demonstrated greater robustness compared to existing approaches.

\section{Related Work}
\label{sec2}

\subsection{Irregularly sampled multivariate time series}
\label{subsec1}

Irregularly sampled time series refer to sequences in which observations are recorded at non-uniform time intervals. When extended to the multivariate setting, this irregularity becomes more complex, as individual variables may be observed at different and inconsistent time points. Consequently, the multivariate time series suffers from asynchronous sampling across variables and a higher rate of missing values. As the number of variables increases, so does the overall sparsity, further complicating downstream modeling.

To address multivariate time series modeling effectively, various architectures such as Long Short-Term Memory (LSTM) \cite{hochreiter1997long} and Transformer \cite{vaswani2017attention} have been proposed. However, these models generally assume regularly sampled time series, where data is collected at uniform intervals, and often require fixed-length sequences as input. 
To satisfy these assumptions, it is common practice to impute missing values and resample the time series at regular intervals. This preprocessing step transforms irregular time series into regular ones at the cost of potentially introducing bias. 

A widely adopted imputation strategy is forward filling \cite{thorsen2020dynamic}, where missing values are replaced with the most recently observed value. To retain information about missingness, binary masks indicating the presence or absence of observations are often appended as auxiliary inputs \cite{kim2024development, choi2024sequential}. Linear interpolation methods \cite{vincent2014assessment, benchekroun2023impact} estimate missing values based on surrounding observations under the assumption of smooth transitions over time. More advanced techniques leverage deep learning to model the imputation process directly from data, such as M-RNN \cite{yoon2018estimating} and BRITS \cite{cao2018brits}. Although these approaches offer improved imputation fidelity, they often require careful tuning of hyperparameters and variable-specific domain knowledge.

To circumvent the limitations of explicit imputation, recent studies have proposed architectures that directly model irregular time series. \textit{GRU-D} \cite{che2018recurrent} integrates a decay mechanism into the gated recurrent unit (GRU) \cite{chung2014empirical} architecture, allowing each variable's missing value imputation to be dynamically performed during training. \textit{SeFT} \cite{horn2020set} treats irregularly sampled time series as a set of tuples (time $t$, observed value $z$, variable indicator $m$) and applies a differentiable set function, leading to favourable results in classification tasks. \textit{mTAND} \cite{shukla2021multi} learns continuous latent embeddings through an attention operation between observed values within an asynchronous time series and predefined reference time points. \textit{IP-Net} \cite{shukla2019interpolation} proposes a semi-parametric interpolation method to obtain regularly sampled time series. \textit{$\textrm{DGM}^2$} \cite{wu2021dynamic} employs a kernel-based method to impute irregular time series by interpolating them with respect to a set of reference points. \textit{Raindrop} \cite{zhang2021graph} constructs a hierarchical temporal graph and utilizes neural message passing to capture cross-variable dependencies, generating high-quality representations for irregularly sampled time series. \textit{ViTST} \cite{li2023time} transforms multivariate time series into a single image using matplotlib and applies a pre-trained vision transformer for classification, demonstrating a simple yet effective approach. While these methods have demonstrated improved performance in modeling irregular time series, they rely solely on numerical imputation or structural interpolation. In contrast, our proposed approach harnesses the LLM’s intrinsic ability to interpret sequential data and contextual cues, enabling semantic reasoning over missing information, thus offering a fundamentally different approach to modeling sparse multivariate time series.

\subsection{Leveraging Large Language Model for time series modeling}
\label{subsec2}
Time series modeling typically requires significant domain expertise and task-specific architectural design, in sharp contrast to recent advances in foundation models for natural language processing (NLP) and computer vision (CV). In these fields, large-scale models are pre-trained on massive datasets and subsequently fine-tuned for various downstream tasks, exhibiting strong generalization capabilities across domains. 
Although similar paradigms are beginning to emerge in time series research \cite{garza2023timegpt, woo2024unified, das2024decoder, goswami2024moment}, their applicability to complex, irregular, multivariate time series classification problems, particularly within domain-specific contexts like healthcare, remains limited and underexplored.
Meanwhile, LLMs such as LLaMA \cite{touvron2023llama, touvron2023llama-2} and GPT-4 \cite{achiam2023gpt} have rapidly advanced, showing exceptional performance across a wide range of interdisciplinary applications \cite{cascella2023evaluating, thirunavukarasu2023large, jablonka202314, vakayil2024rag}. Motivated by these successes, researchers have increasingly sought to leverage LLMs to enhance time series modeling by capitalizing on their powerful pattern recognition and contextual reasoning capabilities.

Recent work on applying LLMs to time series can broadly be categorized into two approaches: direct utilization of pre-trained LLMs, and the development of adaptation mechanisms to bridge the modality gap between structured time series and natural language input.

LLMTime \cite{gruver2023large} proposes a method that converts time series data into text containing numerical values, allowing the LLM to generate natural language-based predictions. This approach enables zero-shot forecasting without additional training, achieving comparable or superior performance compared to traditional time series models (e.g., ARIMA), deep learning-based models (e.g., TCN \cite{lea2017temporal}, N-BEATS \cite{oreshkin2019n}), and Transformer-based models (e.g., Informer \cite{zhou2021informer}, Autoformer \cite{wu2021autoformer}). One Fits All \cite{zhou2023one} feeds time series patches into a pre-trained Transformer and performs forecasting using an output linear layer. This model freezes the parameters of the self-attention layer and feed-forward neural networks, fine-tuning only the layer normalization layers and positional embeddings. The authors attribute the model's ability to effectively handle time series tasks despite freezing most parameters to the universality of the self-attention mechanism. TEMPO \cite{cao2023tempo} decomposes time series into trend, seasonality, and residual components, embeds them into patch representations, and combines each component with an instruction prompt before feeding them into an LLM. These prompts facilitate the learning process of an adaptation layer integrated into the LLM, leading to SoTA performance in time series forecasting benchmarks. TimeLLM \cite{jin2023time}, in contrast, proposes a more direct and parameter-efficient approach: it reprograms time series data into the language space and feeds them into a pre-trained LLM without modifying its core architecture. By combining statistical summaries, task descriptions, and domain knowledge in prompt form, TimeLLM achieves strong performance across diverse forecasting benchmarks—despite training less than 0.2\% of the model’s total parameters.

However, a major limitation of these methods is their assumption of fully observed, regularly sampled time series inputs. In real-world scenarios, particularly in healthcare, time series data are often irregularly sampled and contain substantial missingness. As such, existing LLM-based approaches are not directly applicable without extensive pre-imputation or structural reformatting, both of which may obscure clinically meaningful patterns and introduce unwanted artifacts.
To address this limitation,  we propose VITAL, which extends the reprogramming framework of TimeLLM to irregularly sampled multivariate time series with explicit missingness. The core novelty of VITAL lies in its ability to map sparse EHR time series into a language representation space while preserving both observed values and their missing patterns in a variable-aware manner. Specifically, for vital sign sequences, missing values are encoded using interpretable textual tokens inserted at appropriate time steps, allowing the LLM to reason about the presence or absence of clinical measurements in context. To the best of our knowledge, VITAL is the first model to explicitly encode missingness within irregular multivariate clinical time series using natural language constructs and to leverage a pre-trained LLM for downstream classification tasks without modifying its internal parameters. This approach bridges structured EHR data and unstructured language models in a semantically meaningful and scalable way, enabling contextual reasoning over sparse, heterogeneous clinical sequences.

\section{Methods}
\label{sec3}

\subsection{Problem Formulation}

\begin{figure}[t]
\centering
\includegraphics[width=\textwidth]{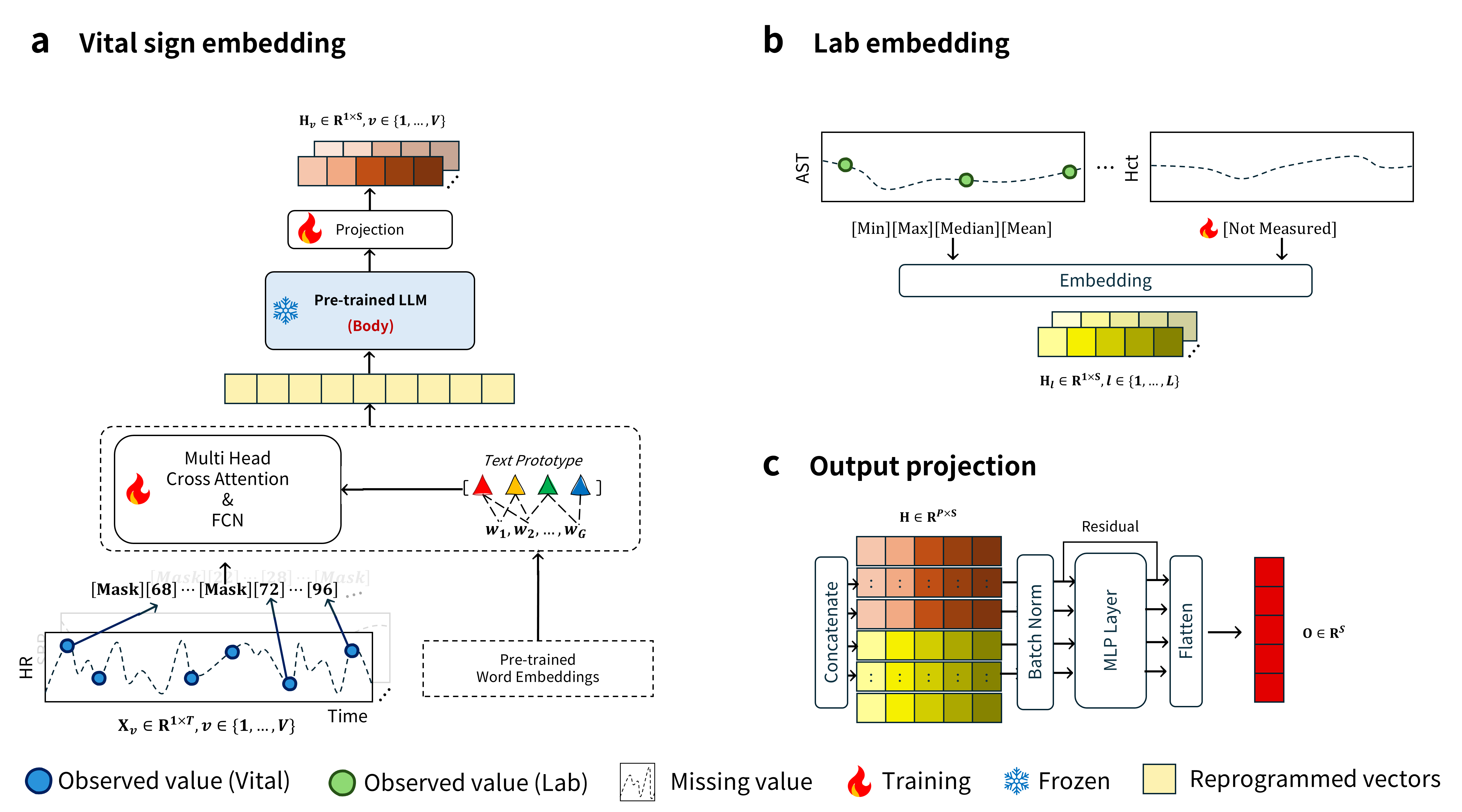}
\caption{An illustration of VITAL. \textbf{a. Vital sign embedding}: Vital signs are reprogrammed into the language modality and fed into a pre-trained LLM. \textbf{b. Lab embedding}: Each lab test is embedded as either a representative value or learnable [Not measured] token, depending on whether it was observed during the observation period. \textbf{c. Output projection}: The vital sign and lab embeddings are concatenated and summarized into fixed-size vector for downstream tasks.}\label{architecture}
\end{figure}

Let $\mathcal{D} = \left\{(S_i, y_i) | i = 1, \ldots, N\right\}$ denote a multivariate time series derived from an EHR database with $N$ labeled samples. Each sample (i.e., patient) corresponds to a label $y_i \in \left\{1,\ldots,C \right\}$, which is one of $C$ classes. Specifically, $S_i$ is a multivariate time series consisting of $P$ variables and spanning $T$ time steps, and can be represented as $S_i = (x_1, x_2, \dots, x_T) \in \mathbb{R}^{T\times P}$. Among the $P$ variables, $V$ represents the number of vital signs, while $L$ denotes the number of lab tests. Here, $x_t^p$ denotes the value of the $p$-th variable at time step $t$. Dataset $\mathcal{D}$ exhibits a substantial degree of missingness since each patient’s measurements were taken irregularly during their stay in the ICU or general ward. As depicted in Figure \ref{architecture}, given an irregularly sampled time series dataset $\mathcal{D}$, the VITAL learns a vital sign embedding function $f: S_{i, v} \rightarrow H_{i, v}$ and a lab test embedding function  $g: S_{i, l} \rightarrow H_{i, l}$, where f and g map  $S_{i, v\,or\,l}$  to a fixed-length vector $H_{i, v\,or\,l}$, respectively. These two vectors, $H_{i, v}$ and $H_{i, l}$, are then concatenated and passed through a mixer layer to obtain the final representation $O_i$. The VITAL then utilizes $O_i$ for classification as well as other downstream tasks of interest.

\subsection{Model architecture}

In this study, we consider the characteristics of different variables to effectively handle a substantial degree of missingness in EHR time series. Since vital signs are automatically measured at the bedside, they generally have fewer missing values compared to other variables, making it meaningful to model their temporal patterns. In contrast, laboratory tests are performed on demand for diagnostic and treatment purposes, resulting in a relatively high proportion of missing values, which diminishes the utility of modeling their temporal patterns. To account for these variable-specific characteristics, we partition the given input time series into vital signs and lab test results and adopt learning strategies tailored to each type of variable. While this strategy is motivated by the structure of EHR data, it can be generalized to broader multivariate time series settings. By leveraging properties such as the missing ratio, variable groups can be flexibly defined. This process is implemented through three steps: \ref{vital} Vital sign embedding (reprogramming vital signs and leveraging a pre-trained large language model), \ref{lab} Lab test result embedding (replacing missing records with representative values or a trainable \textit{Not measured} token), and \ref{mixer} Feature \& Dimension Mixing. 

\subsubsection{Vital sign embedding}
\label{vital}
In this section, we describe the process of transforming vital signs with missing values into temporal contextual representations by leveraging the representation capability of a pre-trained LLM. At the initial stage, vital signs are masked at time points where missing values occur using the masking vector $M = (m_1, m_2, ..., m_T) \in \mathbb{R}^{T\times V}$. The masking vector for each variable is given by

\[
m_t^v = \left\{
\begin{array}{l l}
1,&\text{ if }x_t^v\text{ is observed} \\
0,&\text{ otherwise}
\end{array}
\right.
\]

Vital signs are then subsequently split into $V$ univariate time series. Each univariate time series is then independently fed into the model following the channel independence approach. A univariate time series is represented as $X_{v} \in \mathbb{R}^{1\times T}, v \in \left\{1, \ldots, V\right\}$, to which normalization and reprogramming are applied.

Reprogramming is a technique that introduces learned noise into the target sample, enabling the source model to generate a desired target output without updating its parameters. By leveraging this reprogramming concept, TimeLLM applies reprogramming to time series (i.e., target samples) and feeds them into an LLM (i.e., source model), achieving high predictive performance. This experimentally demonstrates the feasibility of using LLM’s reasoning capabilities for time series data. 
Our adoption of the TimeLLM framework was primarily driven by its parameter-free reprogramming mechanism, which enables the transformation of multivariate time series into input representations compatible with pre-trained LLMs. This approach preserves the model's capabilities for capturing temporal dependencies and contextual relationships, while avoiding the need for fine-tuning or modification of the LLM's internal weights. Such architectural decoupling facilitates the application of powerful pretrained models to temporal clinical data with minimal task-specific adaptation.

In TimeLLM, time series reprogramming is implemented by embedding time series into patches, followed by multi-head cross attention with pre-trained word embeddings $E \in \mathbb{R}^{G \times D} $. However, since the direct correspondence between pre-trained words and specific time series patches is unknown in advance, they apply linear probing to $ E $ to obtain text prototypes $ E' \in \mathbb{R}^{G' \times D} $, where $ G' < G $. This allows the model to adaptively select useful word noise during training.

\begin{figure}[t]
\centering
\includegraphics[width=\textwidth]{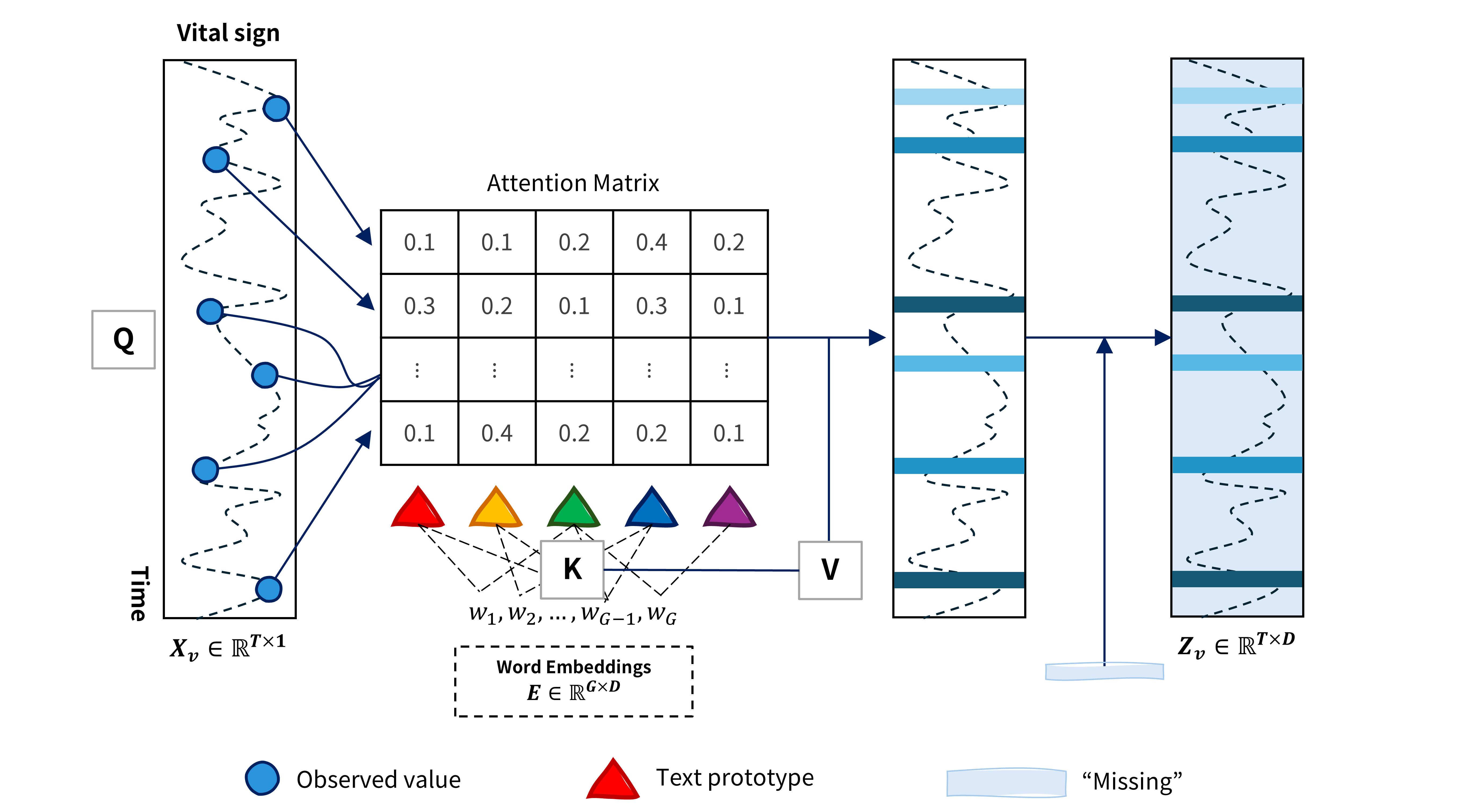}
\caption{Vital sign reprogramming process. Reprogramming is conducted via cross-attention between the observed vital sign value and text prototype. For time points with missing values, the word ``Missing" is embedded.}\label{Reprogramming}
\end{figure}

However, the TimeLLM reprogramming approach exhibits limitations when applied to time series data with missing values. We identify two primary challenges. First, embedding time series with missing values into patches is practically infeasible. Currently, there exists no standardized or principled method for patch-wise embedding of time series with incomplete observations. Naively imputing missing values solely to enable patch construction introduces artificial bias and may distort the underlying temporal dynamics, thereby compromising data integrity. Second, it is semantically and methodologically advantageous to explicitly encode “\textbf{Missing}” as an informative signal within the model input. Given that pre-trained LLMs are adept at extracting contextual meaning from textual sequences, representing missing observations with tokens or structures that convey the concept of “\textbf{Missing}” enables the model to reason about these gaps in a context-aware manner.  This enables more principled utilization of the LLM’s representational capacity by preserving the semantic integrity of missing data, thereby eliminating the need for heuristic imputations or ad hoc structural modifications to the input space.

To address these challenges, we propose a reprogramming approach designed for time series with missing values, as illustrated in Figure \ref{Reprogramming}. In this method, only the observed values are utilized as query inputs in the multi-head cross attention process. Specifically, for each attention head $ k = \left\{1, …, K \right\} $, we define the query matrices as $ Q_k = \operatorname{repeat}(X_{v} \odot m^v, \operatorname{repeat} = d),$ the key matrices as $ K_k = E'W^{key}$, and the value matrices as $V_k = E'W^{value}$, where $ W^{key}, W^{value} \in \mathbb{R}^{D \times d}. $ Here, $D$ represents the hidden dimension of the backbone LLM, and $ d $ denotes the key and value embedding dimensions. The repeat operation in the query ensures dimensional consistency between the key and query for the attention operation. Ultimately, the reprogramming of observed values in each attention head is performed through the following operation:

\begin{equation}
\hat{Z_k} = \textrm{Softmax}(\frac{Q_kK_k}{\sqrt{d_k}})V_k
\end{equation}

After aggregating the computations across each head, we obtain $\hat{Z} \in \mathbb{R}^{T \times d \cdot K} $. Then, to enable direct input into the LLM, $ \hat{Z} $ is linearly projected to match the hidden dimension of the backbone LLM, resulting in a matrix of size $ Z \in \mathbb{R}^{T \times D} $. At the final stage of reprogramming, the word embedding for \textbf{``Missing"} is inserted at time steps where missing values persist, thereby compensating for missing information. Note that the word embeddings used here are identical to those employed during the LLM’s pretraining. Through this process, the vital sign reprogramming is completed.

After feeding the reprogrammed vector $ Z $ into the LLM, we extract the latent embedding of the last time point and apply a fully connected layer to reduce the backbone LLM's hidden dimension to $ S $, obtaining the final vital sign embedding vector $ H_v \in \mathbb{R}^{V \times S} $. We use only the embedding from the final time step because the LLM employed in this study, GPT-2 \cite{radford2019language}, performs inference in an autoregressive manner. As a result, the final time step's embedding encapsulates the contextual information of the entire sequence.
Unlike simple pooling operations, which aggregate features by taking averages or selecting extrema, a fully connected layer enables adaptive feature transformation, preserving richer representations and capturing more nuanced dependencies among latent dimensions.  
This approach allows the model to retain task-relevant information while reducing dimensionality, ensuring that the final embedding maintains essential contextual details.

\subsubsection{Lab test result embedding}
\label{lab}

Laboratory test results collected in the EHR database are recorded far less frequently than vital signs. According to Allyn et al. \cite{allyn2022descriptive}, a retrospective statistical analysis of 138,734 ICU admissions from the eICU database (2014–2015) \cite{pollard2018eicu} revealed that the median time interval between measurements was 10 hours for chemistry parameters (e.g., plasma sodium, potassium, chloride, bicarbonate, urea, and creatinine) and 14 hours for complete blood count parameters (e.g., white blood cell count, red blood cell count, hemoglobin levels, and platelet count). As a result, when laboratory test variables are combined with automatically measured vital signs to construct a multivariate time series, the proportion of missing values per time step increases significantly. 

Unlike vital signs, which exhibit consistent temporal patterns, laboratory test variables do not follow a well-defined temporal structure. In this study, rather than adopting complex time series pattern recognition methods for these variables, we aim to embed key information such as measurement presence and representative values. Figure \ref{architecture}-b illustrates the embedding process for laboratory test results. In this module, the embedding strategy for each variable depends on whether it was measured at least once during the observation period. 

For lab variables that were measured at least once, we extract representative statistics (minimum, maximum, median, and mean values) from the recorded measurements and input them into the embedding layer. This approach enables us to capture key information from lab test results over the observation period effectively. On the other hand, for lab variables that were never measured during the observation period, a single [Not measured] token is used as input to the embedding layer, without distinguishing between variables. The reason for this uniform token assignment is to mitigate potential overfitting, which may arise due to the heterogeneous nature of measured variables across different patients. By keeping the [Not measured] token as a learnable parameter, we allow the model to incorporate missing information in a way that is beneficial for the given prediction task. Through this process, lab test results $ X_l \in \mathbb{R}^{L \times T} $ are transformed into the latent representation $ H_l \in \mathbb{R}^{L \times S} $.

\subsubsection{Feature-Dimension Mixing and Output Projection}
\label{mixer}

Based on the vital sign embedding $ H_v $ and the laboratory test result embedding $ H_l $, we generate the representation vector $ O_i $ for each sample $ S_i $. Instead of simply concatenating $ H_v $ and $ H_l $ followed by flattening, we utilize mixing layers to more effectively capture cross-variate relationships across multiple dimensions (i.e. $ S $ ). While simple flattening preserves independent information for each variable, it has limitations in learning covariate relationships. To address this, we adopt an approach that efficiently extracts both covariate-level and embedding-level interactions by applying feature mixing and dimension mixing. This approach is inspired by the feature \& time mixing layers introduced in TSMixer \cite{chen2023tsmixer}, which enables strong representation learning without relying on conventional RNNs or attention-based models. Notably, while TSMixer \cite{chen2023tsmixer} performs time mixing along the temporal axis, we reinterpret this mechanism to perform mixing along the embedding dimension instead, enabling richer interactions across the representation space of each variable. Subsequently, we incorporate demographic information, which remains constant over time for each patient, into the mixed embedding. This demographic information is concatenated with the mixed embedding, followed by flattening to obtain a final vector of size $ S $. This structure is designed to be applicable to a wide range of downstream tasks.

\section{Implementation consideration}
\label{sec4}

\subsection{Padding reprogrammed time series}
Each ICU patient has a different length of stay, resulting in varying observation times in EHR-based multivariate time series data. To ensure a consistent modeling approach, VITAL applies padding to the reprogrammed time series when a patient's length of stay is shorter than the predefined maximum duration. In this case, padding is added before the admission time rather than after the final recorded time step. This is to ensure that we don't utilize the output corresponding to the padded data as vital sign embedding. 

\subsection{Variable Partitioning Strategy} 
We partition the input time series into vital signs and laboratory test variables to reflect their distinct characteristics. However, exploratory data analysis on the training data revealed that some vital sign variables exhibited extremely high missing ratios, comparable to those of laboratory test variables. For instance, in the P19 \cite{reyna2020early} dataset, temperature and EtCO2(end tidal carbon dioxide) had missing rates of 66.31\% and 97.12\%, respectively. Similarly, in the P12 \cite{goldberger2000physiobank} dataset, MAP(mean arterial pressure), NISBP(non-invasive systolic arterial blood pressure), and RR(respiration rate) had missing rates of 89.66\%, 95.38\%, and 97.29\%, respectively. Based on these observations, we treated these variables as lab-like and processed them using the lab test embedding module. This flexible partitioning strategy reflects the principle that the modeling method should be selected based on the data characteristics (e.g., missingness), not just the source of measurement.

\subsection{Hyperparameter}
All experiments were conducted using an NVIDIA A6000 GPUs with 48G memory. Additionally, the learning rate was set to 0.001, and the batch size was consistently maintained at 128 across all datasets to ensure experimental consistency. The implementation code of the proposed model is available at \href{https://github.com/Jeong-Eul/VITAL/tree/main}{https://github.com/Jeong-Eul/VITAL}.
    
\section{Experiment}
\label{sec5}

To demonstrate the effectiveness of VITAL, we evaluate its classification performance and robustness in Sections \ref{sec5.2} and \ref{sec5.3}, respectively. To enhance interpretability and provide transparency into the model's internal mechanisms, we present quantitative evidence in Section \ref{sec5.5} by analyzing performance changes under different missing value handling strategies. Furthermore, qualitative insights are offered through visualizations in Sections \ref{sec5.4} and \ref{sec5.6}.

\begin{table*}[htbp]
\caption{Dataset information. `Vars' refers to variables, `Demo' represents demographic information, and `MR' refers to missing ratio.
} 
\vspace{2mm}
\centering
\small
\resizebox{\textwidth}{!}{
\begin{tabular}{lcccccccc}
\toprule
Datasets & Patients & Vars & Classes & Demo & Class ratio (Neg : Pos) & MR (total) & MR (vital) & MR (lab)\\ \midrule
P19 & 38,803 & 34 & 2 & True & 0.96:0.04 &94.9\% &32.9\% &95.1\%\\
P12 & 11,988 & 36 & 2 & True & 0.86:0.14 &88.4\% &70\% &81.4\%\\
\bottomrule
\end{tabular}
}
\label{tab:datasets}
\end{table*}

\subsection{Details on dataset and metrics}
As summarized in Table \ref{tab:datasets}, this study utilizes two widely used ICU benchmark datasets. The first dataset, PhysioNet Sepsis Early Prediction Challenge 2019 (P19) \cite{reyna2020early}, contains data from 38,803 patients, including 34 variables and binary labels indicating sepsis onset. The P19 \cite{reyna2020early} is used for a binary classification task that predicts whether a patient will develop sepsis based on a multivariate irregular time series recorded over 60 hours. The second dataset, PhysioNet Mortality Prediction Challenge 2012 (P12) \cite{goldberger2000physiobank} contains data from 11,988 patients, including 36 variables and binary labels indicating in-hospital mortality. The P12 is used for a binary classification task that predicts patient mortality based on a multivariate irregular time series recorded over 215 hours. For this study, we utilized the preprocessed data provided by Raindrop \cite{zhang2021graph}. To ensure fair comparisons with prior studies, we adopted the same data split used in all baseline models. Since both datasets exhibit label imbalance, we employed AUROC (Area Under the Receiver Operating Characteristic Curve) and AUPRC (Area Under the Precision-Recall Curve) as evaluation metrics.

\subsection{Time series classification}
\label{sec5.2}
\begin{table*}[!t]
\scriptsize
\centering
\caption{Evaluation of baseline models on irregular time series classification. Models with the highest and second-highest performance are denoted by bold and \underline{underline}, respectively.}
\vspace{2mm}
\label{tab:main_result}
\resizebox{\textwidth}{!}{ 
\begin{tabular}{l|ll|ll}
\toprule
& \multicolumn{2}{c|}{P19} & \multicolumn{2}{c}{P12} \\ \cmidrule{2-5}
\multirow{-2}{*}{Methods} & AUROC & AUPRC & AUROC & AUPRC \\ \midrule
Transformer & \valstd{80.7}{3.8} & \valstd{42.7}{7.7} & \valstd{83.3}{0.7} & \valstd{47.9}{3.6} \\
Trans-mean & \valstd{83.7}{1.8} & \valstd{45.8}{3.2} & \valstd{82.6}{2.0} & \valstd{46.3}{4.0} \\
GRU-D & \valstd{83.9}{1.7} & \valstd{46.9}{2.1} & \valstd{81.9}{2.1} & \valstd{46.1}{4.7} \\
SeFT & \valstd{81.2}{2.3} & \valstd{41.9}{3.1} & \valstd{73.9}{2.5} & \valstd{31.1}{4.1} \\
mTAND & \valstd{84.4}{1.3} & \valstd{50.6}{2.0} & \valstd{84.2}{0.8} & \valstd{48.2}{3.4} \\
IP-Net & \valstd{84.6}{1.3} & \valstd{38.1}{3.7} & \valstd{82.6}{1.4} & \valstd{47.6}{3.1} \\
DGM$^2$-O & \valstd{86.7}{3.4} & \valstd{44.7}{11.7} & \valstd{84.4}{1.6} & \valstd{47.3}{3.6} \\
MTGNN & \valstd{81.9}{6.2}  & \valstd{39.9}{8.9} & \valstd{74.4}{6.7} & \valstd{35.5}{6.0} \\
Raindrop & \valstd{87.0}{2.3} & \valstd{51.8}{5.5} & \valstd{82.8}{1.7} & \valstd{44.0}{3.0} \\
ViTST & \valstdu{89.2}{2.0} & \valstdu{53.1}{3.4} & \valstdu{85.1}{0.8}  & \valstdu{51.1}{4.1} \\
\midrule
\textbf{VITAL} & \valstdb{89.3}{2.1} & \valstdb{57.5}{3.1} & \valstdb{86.0}{1.4} & \valstdb{54.9}{4.2} \\
\bottomrule
\end{tabular}
}
\end{table*}

We evaluated the performance of VITAL by comparing it with existing models designed for irregularly sampled time series. The comparison included Transformer \cite{vaswani2017attention}, Trans-mean (a version trained with an imputation method that replaces missing values with the mean of the observed values for each variable), GRU-D \cite{che2018recurrent}, SeFT \cite{horn2020set}, mTAND \cite{shukla2021multi}, IP-Net \cite{shukla2019interpolation}, Raindrop \cite{zhang2021graph}, and the SoTA model ViTST \cite{li2023time}. To ensure a fair comparison and avoid discrepancies caused by differences in the experimental environment, we did not re-evaluate the baseline models. Instead, we directly adopted the reported performance of baselines from the ViTST paper \cite{li2023time} for comparison with VITAL.  As shown in Table \ref{tab:main_result}, VITAL achieved superior performance over existing baseline models. Compared to the previous SoTA model, ViTST, VITAL improved AUROC and AUPRC by 0.1\% and 8.2\% on P19 \cite{reyna2020early}, and by 1.1\% and 6.6\% on P12 \cite{goldberger2000physiobank}, respectively, demonstrating its effectiveness in handling imbalanced data.

\subsection{Leave-fixed-sensors-out}
\label{sec5.3}

\begin{table}[!htb]
\centering
\caption{Classification on samples with fixed missing sensors (P19)}
\label{tab:setting2_P19}
\resizebox{\textwidth}{!}{
\begin{tabular}{l|ll|ll|ll|ll|ll}
\toprule
\multirow{2}{*}{Models} & \multicolumn{10}{c}{Missing ratio} \\ \cmidrule{2-11}
 & \multicolumn{2}{c|}{10\%} & \multicolumn{2}{c|}{20\%} & \multicolumn{2}{c|}{30\%} & \multicolumn{2}{c|}{40\%} & \multicolumn{2}{c}{50\%} \\ \cmidrule{2-11}
 & AUROC & AUPRC & AUROC & AUPRC & AUROC & AUPRC & AUROC & AUPRC & AUROC & AUPRC \\ \midrule
Transformer & 77.4 $\pm$ 3.5 & 38.2 $\pm$ 4.2 & 75.7 $\pm$ 3.4 & 35.2 $\pm$ 5.4 & 75.1 $\pm$ 3.5 & 35.5 $\pm$ 4.4 & 75.3 $\pm$ 3.5 & 36.2 $\pm$ 4.2 & 74.9 $\pm$ 3.1 & 35.5 $\pm$ 5.0 \\
Trans-mean & 79.2 $\pm$ 2.7 & 40.6 $\pm$ 5.7 & 79.8 $\pm$ 2.5 & 38.3 $\pm$ 2.8 & 76.9 $\pm$ 2.4 & 37.5 $\pm$ 5.9 & 76.4 $\pm$ 2.0 & 36.3 $\pm$ 5.8 & 74.1 $\pm$ 2.3 & 41.3 $\pm$ 4.7 \\
GRU-D & 79.6 $\pm$ 2.2 & 37.4 $\pm$ 2.5 & 77.5 $\pm$ 3.1 & 36.5 $\pm$ 4.6 & 76.6 $\pm$ 2.9 & 35.1 $\pm$ 2.4 & 74.6 $\pm$ 2.7 & 35.9 $\pm$ 2.7 & 74.1 $\pm$ 2.9 & 33.2 $\pm$ 3.8 \\
SeFT & 77.3 $\pm$ 2.4 & 25.5 $\pm$ 2.3 & 63.5 $\pm$ 2.0 & 14.0 $\pm$ 1.1 & 62.3 $\pm$ 2.1 & 12.9 $\pm$ 1.2 & 57.8 $\pm$ 1.7 & 9.8 $\pm$ 1.1 & 56.0 $\pm$ 3.1 & 7.8 $\pm$ 1.3 \\
mTAND & 79.7 $\pm$ 2.2 & 29.0 $\pm$ 4.3 & 77.8 $\pm$ 1.9 & 25.3 $\pm$ 2.4 & 77.7 $\pm$ 1.9 & 27.8 $\pm$ 2.6 & 79.4 $\pm$ 2.0 & 32.1 $\pm$ 2.1 & 77.3 $\pm$ 2.1 & 27.0 $\pm$ 2.5 \\
Raindrop & 84.3 $\pm$ 2.5 & 46.1 $\pm$ 3.5 & 81.9 $\pm$ 2.1 & 45.2 $\pm$ 6.4 & 81.4 $\pm$ 2.1 & 43.7 $\pm$ 7.2 & \underline{81.8 $\pm$ 2.2} & 44.9 $\pm$ 6.6 & 79.7 $\pm$ 1.9 & 43.8 $\pm$ 5.6 \\ 
VITAL & \textbf{87.8 $\pm$ 1.7} & \underline{53.2 $\pm$ 3.9} & \underline{84.6 $\pm$ 1.9} & \underline{48.7 $\pm$ 3.6} & \underline{82.7 $\pm$ 1.9} & \underline{47.5 $\pm$ 3.9} & 81.3 $\pm$ 2.0 & \underline{46.3 $\pm$ 4.1} & \underline{80.6 $\pm$ 1.6} & \underline{45.3 $\pm$ 3.9} \\
VITAL (lab) & \underline{87.6 $\pm$ 2.0} & \textbf{54.0 $\pm$ 2.4} & \textbf{86.6 $\pm$ 2.0} & \textbf{52.5 $\pm$ 2.0} & \textbf{86.7 $\pm$ 2.1} & \textbf{52.5 $\pm$ 1.9} & \textbf{86.0 $\pm$ 2.0} & \textbf{51.8 $\pm$ 1.4} & \textbf{85.9 $\pm$ 2.3} & \textbf{52.4 $\pm$ 2.2} \\
\bottomrule
\end{tabular}
}
\end{table}

We further evaluated whether VITAL could maintain high performance even when specific variables were entirely absent using P19 \cite{reyna2020early} dataset. This setting simulates real-world clinical scenarios where decisions must be made despite missing critical variables, allowing us to assess the model’s robustness under such conditions. In this experiment, we selected a subset of variables and set them as completely missing in the test sets. We then applied the trained model to evaluate its performance. For a fair comparison, we followed the exact variable removal order defined in the \textit{leave-fixed-sensors-out} setting of Raindrop \cite{zhang2021graph}. This order was determined by sorting the variables based on their information gain from a random forest classifier, with all vital signs ranked within the top 20\%.

The results are presented in Table \ref{tab:setting2_P19}, where we observe that VITAL outperforms all baseline models except for Raindrop \cite{zhang2021graph} in AUROC when the missing ratio reaches 40\%. In this experimental setting, a missing ratio of 20\% corresponds to the complete removal of all vital signs used for modeling, meaning that the LLM receives only the word ``Missing" at all time points. As a result, the generated vital sign embedding vector lacks any contextual information from the time series. Despite this, the rate of performance degradation as the missing ratio increased remained comparable to that of Raindrop \cite{zhang2021graph}. To further simulate real ICU environments, we conducted an additional experiment in which only laboratory test variables were removed (VITAL (lab) in Table \ref{tab:setting2_P19}). This setup reflects a realistic clinical scenario, as vital signs are rarely missing in practice. The results indicate that when only laboratory test variables were removed, the model’s performance degraded at a significantly slower rate compared to when both laboratory tests and vital signs were removed. Notably, even after eliminating half of the laboratory test variables, VITAL achieved a higher AUPRC than the baseline model of Raindrop \cite{zhang2021graph}. These findings highlight the robustness of VITAL in handling missing variables and its potential applicability in real-world clinical settings.

\subsection{Ablation study on [Not measured] token}
\label{sec5.4}

\begin{table*}[!htb]
\scriptsize
\centering
\caption{Differences in classification performance by training strategy.}
\label{nm_strategy}
\resizebox{\textwidth}{!}{ 
\begin{tabular}{l|cc|cc}
\toprule
& \multicolumn{2}{c|}{P19} & \multicolumn{2}{c}{P12} \\ \cmidrule{2-5}
\multirow{-2}{*}{Strategy} & AUROC & AUPRC & AUROC & AUPRC \\ \midrule
Trainable & \valstd{89.3}{2.1} & \valstd{57.5}{3.1} & \valstd{86.0}{1.4} & \valstd{54.9}{4.2} \\ \midrule

Zero & \valstd{88.5}{2.1} & \valstd{55.2}{4.0} & \valstd{85.7}{1.5} & \valstd{53.3}{3.6} \\

Random & \valstd{86.0}{1.5} & \valstd{50.5}{3.4} & \valstd{85.3}{0.9} & \valstd{52.1}{2.7} \\ 

\bottomrule
\end{tabular}
}
\end{table*}

In the proposed model VITAL, missing laboratory test results that were never measured throughout the observation period are handled by introducing a learnable [Not measured] token. To evaluate the effectiveness of this strategy, we conducted an ablation study. Specifically, for both the P19 \cite{reyna2020early} and P12 \cite{goldberger2000physiobank} datasets, we compared model performance under three different settings where the [Not measured] token was: (1) embedded as a fixed zero vector, (2) embedded as a random vector, and (3) implemented as a learnable parameter (our proposed method).

The results are summarized in Table \ref{nm_strategy}. Across both datasets, the random setting yielded the lowest performance. Even under this drastic setting, the performance did not fall significantly short of that of the strong baseline model Raindrop \cite{zhang2021graph}. Notably, in the P12 \cite{goldberger2000physiobank} dataset, replacing the token with a zero vector still outperformed the previous SoTA model ViTST. These results highlight that the vital sign embedding module in VITAL contributes significantly to the model’s overall performance. Since we applied standard scaling to the input data, embedding the token as a zero vector corresponds to using the mean value of the variable. This explains why zero embedding generally outperformed random embeddings in both datasets.

\begin{figure}[t]
\centering
\includegraphics{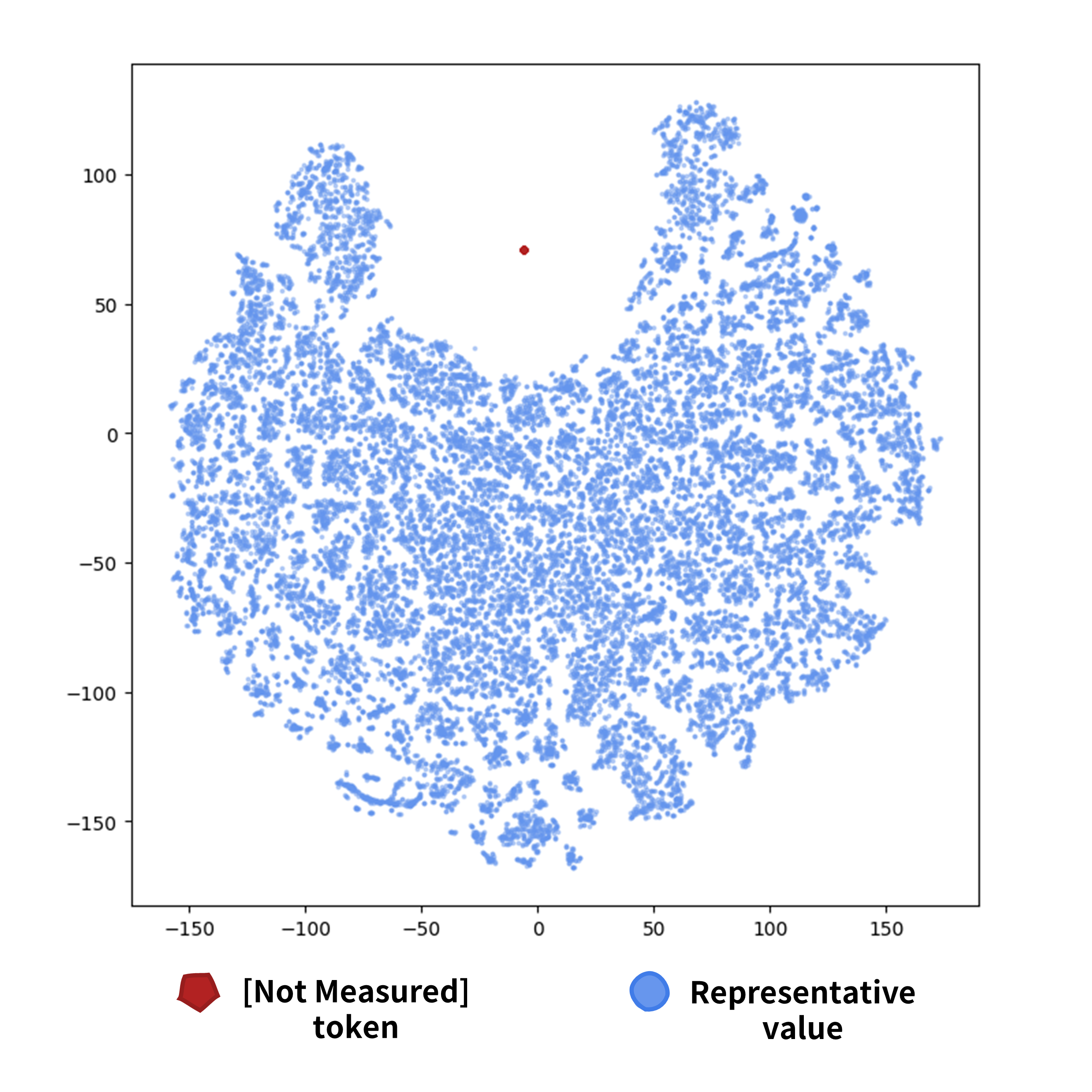}
\caption{Visualization of representative value and [Not measured] token embedding in latent space.}\label{not_measured_token}
\end{figure}

Unlike the [CLS] token in Transformer \cite{vaswani2017attention}, which learns task-relevant representations through attention with surrounding sequences and can be interpreted post hoc, the [Not measured] token in VITAL is simply concatenated with other vectors and linearly projected, making its internal representation difficult to interpret directly. To address this, we analyzed the embedding space by comparing the learned [Not measured] token with representative value embeddings. As visualized in Figure \ref{not_measured_token}, the learned token exhibited a distinct position in the embedding space, clearly separated from those representing observed values. This suggests that the model successfully learned to encode the unique semantics of ``never measured throughout the entire observation period," allowing it to effectively convey this information to downstream components.

\subsection{What if we choose a different word during reprogramming step?}
\label{sec5.5}
\begin{table*}[h]
\scriptsize
\centering
\caption{Ablation studies on missing value reprogramming strategy
}
\label{tab:ablation_1}
\resizebox{\textwidth}{!}{ 
\begin{tabular}{l|cc|cc}
\toprule
& \multicolumn{2}{c|}{P19} & \multicolumn{2}{c}{P12} \\ \cmidrule{2-5}
\multirow{-2}{*}{Variant} & AUROC & AUPRC & AUROC & AUPRC \\ \midrule
Missing & \valstd{89.3}{2.1} & \valstd{57.5}{3.1} & \valstd{86.0}{1.4} & \valstd{54.9}{4.2} \\ \midrule

\textit{Missing} $\rightarrow$ \textit{Null} & \valstd{87.2}{1.9} & \valstd{54.0}{3.2} & \valstd{75.4}{3.4} & \valstd{35.2}{5.8} \\

\textit{Missing} $\rightarrow$ \textit{Apple} & \valstd{86.1}{3.8} & \valstd{51.6}{7.1} & \valstd{72.6}{2.4} & \valstd{30.2}{3.6} \\ 

\textit{Missing} $\rightarrow$ \textit{Engineering} & \valstd{84.3}{3.2} & \valstd{50.3}{3.3} & \valstd{72.4}{3.0} & \valstd{29.6}{2.9} \\ 
\bottomrule
\end{tabular}
}
\end{table*}

The proposed model, VITAL, inserts a word that the LLM can explicitly interpret at each time point where a missing value occurs during the process of reprogramming irregular vital sign sequences into the language space. We hypothesized that this approach enables the LLM to effectively learn contextual representations of sequences that contain missing values. In this section, we evaluate whether the LLM indeed generates meaningful contextual embeddings from such incomplete sequences. To do so, we observe performance changes when replacing the word ``Missing” (previously used at missing time points) with alternative words. The results are summarized in Table \ref{tab:ablation_1}.

When the word ``Missing” was replaced with ``Null,” which also implies absence, the model performance decreased compared to using ``missing.” However, it still outperformed cases where unrelated words such as ``Apple” or ``Engineering” were used. This suggests that ``Null” retains some semantic meaning of absence in the embedding space, allowing the LLM to interpret it more appropriately. Nevertheless, using the word ``Missing” consistently yielded the best performance, indicating that the LLM interprets this token as a clear signal of missingness, and is thereby able to generate high-quality contextual representations of time series with missing data.

\subsection{Understanding the behavior of vital sign embedding step}
\label{sec5.6}
\begin{figure}[h]
\centering
\includegraphics[width=\textwidth]{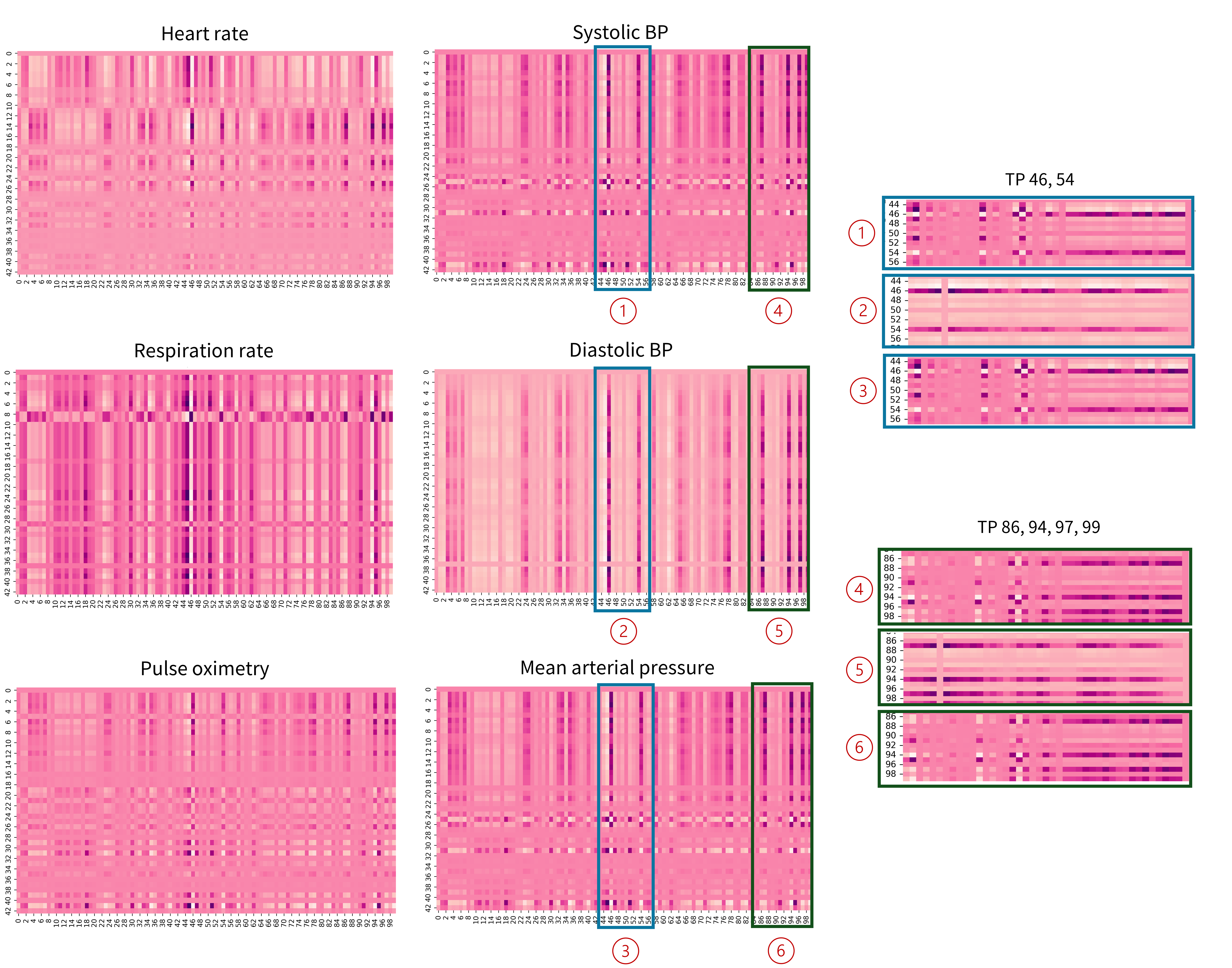}
\caption{Attention scores across variables during the reprogramming process, where the x-axis corresponds to text prototypes and the y-axis to time.}\label{Reprogramming_variable}
\end{figure}

We conducted a case study using the P19 dataset \cite{reyna2020early} to analyze the behavior of the vital sign embedding module. Specifically, we examined how each variable's time series is reprogrammed using 100 text prototypes, as illustrated in Figure \ref{Reprogramming_variable}. Each heatmap visualizes the attention scores of text prototypes (x-axis) over time (y-axis), with darker colors indicating higher attention scores. The visualization revealed that different variables engaged distinct sets of text prototypes during reprogramming. Given that our reprogramming strategy directly uses the observed values as queries, this divergence can be interpreted as a result of differences in the value distributions across variables, which in turn lead to variations in the distribution of queries. In other words, the proposed VITAL preserves the distinct statistical characteristics and temporal patterns of each univariate time series, resulting in reprogrammed vectors that are capable of capturing variable-specific semantics. For example, Systolic BP, Diastolic BP, and Mean arterial pressure are known to be closely related physiological measurements \cite{chemla2005mean, taylor2011impact} (Mean arterial pressure is computed as a weighted average of Systolic and Diastolic BP). These three variables frequently attended to the same set of text prototypes, specifically prototypes 46, 54, 86, 94, 97, and 99, all of which showed consistently high attention scores. These findings suggest that our reprogramming approach not only learns natural language noises, but also effectively reflects the underlying distribution of observed values, thereby producing independent vector representations that preserve the semantic distinctions among variables.

\begin{figure}[h]
\centering
\includegraphics[width=\textwidth]{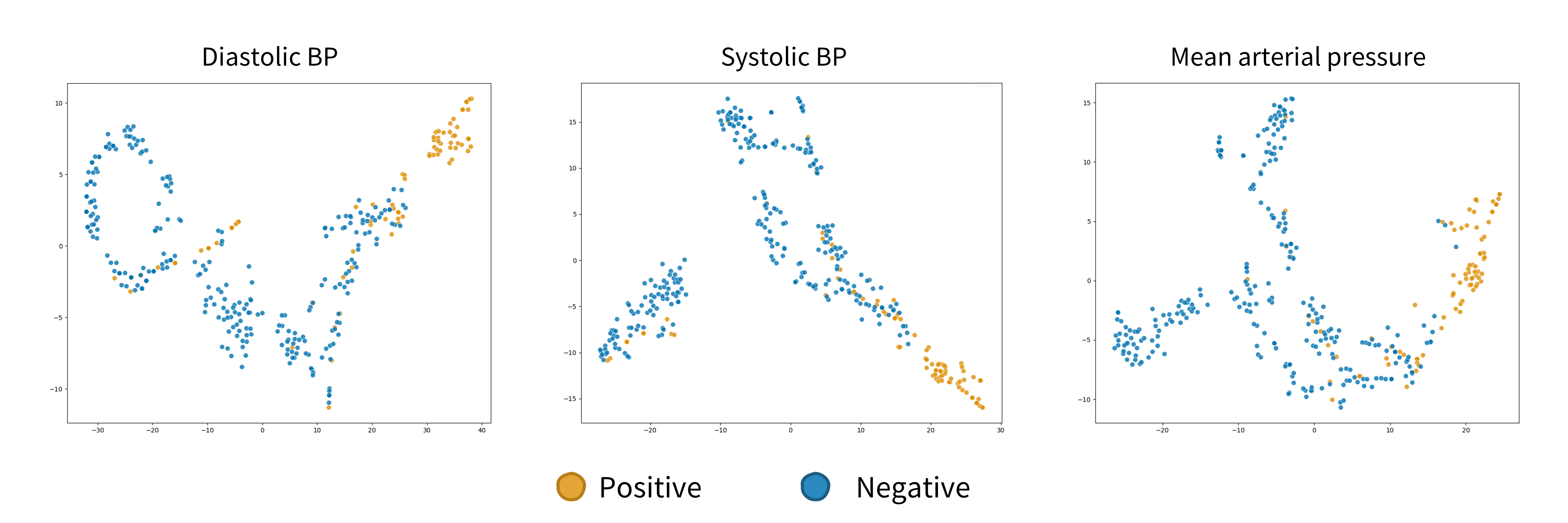}
\caption{Visualization of embeddings of key vital sign variables across different labels, where each vector point corresponds to an individual patient.}\label{vital_embedding}
\end{figure}

We also evaluated whether the embeddings of key vital sign variables differ significantly according to the label. As shown in Figure \ref{vital_embedding}, we visualized the embedding vectors of each patient's Diastolic BP, Systolic BP, and Mean arterial pressure in the latent space. Each vector point corresponds to an individual patient, with blue indicating a negative label (no occurrence of sepsis) and orange indicating a positive label (occurrence of sepsis), both within the next 6 hours. The visualization results show that most patient embeddings exhibit clear separability based on the label, suggesting that the pre-trained LLM effectively captures temporal contextual patterns within the reprogrammed vital signs and reflects them as meaningful features for label prediction.

\section{Limitation and Future work}

Despite its strong empirical performance and efforts to interpret, VITAL presents notable limitations in terms of feature importance interpretability, primarily due to the model's architectural design. The vital sign embedding module builds on the structure of TimeLLM \cite{jin2023time}, which inherently treats each variable as channel-independent. This design prevents the model from explicitly capturing inter-variable dependencies. Similarly, the lab test embedding module aims to extract simple summary statistics or represent missingness through independent embeddings, thereby limiting its ability to model interactions among variables. Furthermore, the final output projection step concatenates all variable-level embeddings and maps them into a single vector via a TSMixer \cite{chen2023tsmixer} structure. Because the resulting input tensor has three dimensions (i.e., [batch size, input features, embedding dimension]), it becomes challenging to directly estimate each variable’s contribution using standard attribution methods such as Integrated Gradients \cite{sundararajan2017axiomatic} or SHAP \cite{lundberg2017unified}, which typically require 2D input tensors (e.g., [batch size, input features]). While we attempted to incorporate an attention mechanism into the output projection layer to enhance feature importance interpretability, this modification led to a noticeable performance drop compared to the use of TSMixer \cite{chen2023tsmixer} structure. Therefore, a key limitation of VITAL lies in its inability to offer variable-level interpretability. Future research should explore ways to better balance interpretability and predictive performance by designing model architectures that are more transparent and allow for easier tracking of how each variable contributes to the prediction. Additionally, further validation is required to ensure the model’s generalizability for clinical deployment, particularly when applied to multi-institutional settings and a broader spectrum of clinical variables.

\section{Conclusion}

In this study, we propose VITAL, a novel tool designed for learning from irregularly sampled physiological time series in EHRs. To the best of our knowledge, this is the first approach to leverage a large language model to such data by explicitly encoding missingness as contextual temporal information. VITAL implements a variable-aware representation strategy that considers the collection mechanisms of clinical variables commonly found in EHRs. For variables such as vital signs which are frequently measured and exhibit clear temporal patterns, VITAL reprograms their sequences into the language space and extracts temporal context-aware embeddings using an LLM. In contrast, for laboratory test variables that are measured infrequently and exhibit high missingness, VITAL embeds representative values or a learnable [Not measured] token depending on whether the variable was observed during the measurement period. This approach distinguishes itself from conventional EHR time series modeling methods by applying customized representation strategies tailored to the inherent characteristics of each variable. Extensive experiments demonstrate that VITAL outperforms state-of-the-art models specifically designed for irregular time series. It achieves superior performance on benchmark EHR datasets and maintains robustness under increasing levels of missing observations. We believe that this work provides both practical contributions and conceptual insights, highlighting a promising direction for future research in medical AI, particularly in the development of early warning systems and foundation models capable of capturing complex, irregular time series data. Moreover, we anticipate that it will foster further investigation into reprogramming-based methodologies for irregular time series across a wide range of domains beyond healthcare.

\bibliographystyle{elsarticle-num} 
\bibliography{reference}

\end{document}